\documentclass{article}

\usepackage{PRIMEarxiv}

\usepackage[utf8]{inputenc} 
\usepackage[T1]{fontenc}    
\usepackage{url}            
\usepackage{booktabs}       
\usepackage{amsfonts}       
\usepackage{nicefrac}       
\usepackage{microtype}      
\usepackage{amsmath}
\usepackage{amsthm}
\usepackage{textcomp}
\usepackage{amssymb}
\usepackage{siunitx}
\usepackage{enumerate}   
\usepackage[shortlabels]{enumitem}
\usepackage{fancyhdr}       
\usepackage{graphicx}       
\graphicspath{{media/}}     

\title{Analysis of business process automation as linear time-invariant system network
}

\author{
  Mauricio Jacobo-Romero$^1$,  Danilo S. Carvalho$^1$,  Andr\'e Freitas$^{1,2}$ \\
  ${}^1$Department of Computer Science, The University of Manchester, Manchester, UK\\
  \texttt{\{mauricio.jacoboromero, danilo.carvalho, andre.freitas\}@manchester.ac.uk} \\
  ${}^2$Reasoning \& Explainable AI group, Idiap Research Institute, Martigny, Switzerland \\
  \texttt{andre.freitas@idiap.ch}
}

\begin{document}
\maketitle

\begin{abstract}
In this work, we examined Business Process (BP) production as a signal; this novel approach explores a BP workflow as a linear time-invariant (LTI) system. We analysed BP productivity in the frequency domain; this standpoint examines how labour and capital act as BP input signals and how their fundamental frequencies affect BP production. Our research also proposes a simulation framework of a BP in the frequency domain for estimating productivity gains due to the introduction of automation steps. Our ultimate goal was to supply evidence to address Solow’s Paradox.
\end{abstract}


\section{Introduction}
A signal is a set of data or information about the nature of a physical phenomenon \cite{lathi1998signal}. Signals frequently exhibit temporal, spatial, or both-attribute relationships. However, solely space changes do not qualify as signals. In other words, the subject of study should vary on time \cite{alma9927799804401631}. To study their properties, we represent signals with mathematical functions with one or more variables in continuous or discrete domains \cite{Chakravorty2018}. 

Production possesses signal characteristics. It evolves its value over time and is a function of various time-dependent inputs \cite{Prokopenko1987}.   From all signal features, frequency is the main object of our research. Frequency is the number of instances of one event per unit of time  \cite{Chakravorty2018}. If we consider a "widget" as the unit of the productive process output, we can say that the productivity definition fits into the frequency concept. Thus, a change in production frequency is equivalent to a change in productivity.

In this fashion, it is possible to analyse productivity with the signal processing theory. Signal processing theory (SPT) analyses signals to alter their properties to optimise their broadcasting, amend distortions due to noise interference, and store them \cite{alma9927799804401631}. 
SPT allowed us to represent a BP as a Linear Time-Invariant (LTI) system. An LTI system is a set of elements that work together as parts of a mechanism. Its main characteristics are time-invariance and linearity. It produces an output signal from an input signal \cite{Ogata2010}. LTI properties permit the depiction of BP productivity as the superposition of all its internal components or tasks. We can analyse, then, each activity as an LTI. Thus, we can escalate down or up our approach.

Our central motivation was to offer evidence to solve Solow's Paradox by examining production as a signal. We proposed to study BP productivity in the frequency domain, explored labour and capital as BP input signals, and their relationship with BP's productivity fundamental frequency. This model is the foundation of our simulation framework.

With this approach, we sought to: (i) provide a granular estimation of productivity gains within a BP workflow, aiming to supply evidence to address Solow's Paradox; (ii) develop an LTI BP-task model to analyse productivity variations, in the frequency domain, after the introduction of automation steps; (iii) integrate automation metrics into the LTI BP-task model to dissect the productivity interventions; (iv) prospectively estimate impacts in productivity due to production shifts; (v) and identify a BP-task resonant "productivity state" - A frequency value that may cause instability on productivity levels.

The paper is organised as follows. Section II presents an overview of the related work. In section III, we introduce key modelling concepts. Our methodology is described in Section IV. We explain our model in section V. Section VI and VII introduce results and discussion, respectively. Finally,we present our conclusions.

\section{Related Work}
In the 1980s, Robert Solow, Nobel-Prize-winning economist, remarked: "we see the computer age everywhere except in the productivity statistics"\cite{ByRobertM.Solow1987WBWO}. This statement is known as "Solow's paradox". His purpose was to exhibit a contradiction between automation's general perception and a discernible reality \cite{brynjolfsson2018artificial}. 

The research community proposed four explanations to elucidate the paradox: (1) false hopes, (2) productivity mismeasurement, (3) concentrated distribution and rent dissipation, and (4) implementation and restructuring lags\cite{brynjolfsson2018artificial}. In 1996, a sample of companies provided data that concluded that Solow's Paradox was no longer happening. The study pointed out loose information as the root cause of this productivity perception \cite{Brynjolfsson1996}. 

In recent years, firms have extensively integrated automation into their business process \cite{VanderAalst2012}. As in the 1980s, companies see no vast improvement in their productivity charts \cite{Denning2021}.

Productivity assessment is challenging due to complex business operations \cite{Martin2014}. To disentangle this intricacy, Business Process Management (BPM) captures, understands, and improves work within an organisation through process models \cite{Martin2016}. BPM's outcome is a series of business processes (BPs). BPs are networks of activities that connect employees and activities across different locations with the help of broad IT systems \cite{Smirnov2012}.

Typically, BPs produce logs that record their operation. Knowledge extraction is the goal of the process mining (PM) field \cite{LianjunAn}. PM provides information to simulation systems. Business process simulation (BPS) is a technique that provides insights into execution times, resource allocations, and costs \cite{Pufahl2018}. BPS objectives target mostly activity forecasting. The ultimate goal is to accurately reconstruct the BP from logs \cite{Jadric2020, Camargo2021}. Recent BPS techniques include machine learning algorithms. Predictive business process monitoring (PBMP) focuses on improving operational efficiency and avoiding deadlocks by its forecasts \cite{DBLP:journals/corr/abs-2104-00721}. PBMP also includes ProcessTransformers to anticipate event time and process completion time \cite{Wickramanayake2022}. However, current PBMP approaches do not overpass dedicated algorithms for these tasks \cite{RiveraLazo2022, Chen2022}. 

On the other hand, PM integrated eXplainable AI (XAI) techniques to identify the relationships between events \cite{Velioglu2022}. However, it is challenging to escalate these results. Therefore, XAI outcomes are significantly limited \cite{Portolani2022, chaima5road}.

ML models deliver more accurate results than Data-Driven simulations (DDS). Nevertheless, ML systems do not allow changes in the process and simulate the consequences of these modifications \cite{Camargo2022}.

Recapitulating, ML techniques are suitable for foreshadowing problems before their occurrence. In other words, they recommend resource reallocation without including productivity assessments \cite{Rama-Maneiro2022, https://doi.org/10.48550/arxiv.1602.04938}. 

PBMP has also integrated signal processing theory (SPT) to extract information from PM logs. SPT deals with cyclical behaviour identification to fine-tune BPS models \cite{RePEc:men:wpaper:07_2011}. Time and Frequency analysis provide underlying cyclical information to predict inflation, production, or resource consumption variations \cite{Bogalo2021}.  

Some approaches combine SPT and graphs to grasp process knowledge. Data low-pass graph filters capture social systems movements. Therefore, these algorithms predict the next steps due to human interactions within the BP \cite{Ramakrishna2020} \cite{Mao2020}.

Other approaches integrate SPT to develop models for social learning. The objective is to explain how knowledge emerges or fails to materialise from process execution. This technique considers employees as social agents. SPT identifies agent trajectories to predict the next interaction based on the environment dynamics \cite{Chamley2013}.

Nevertheless, SPT models work with constrained hypotheses on physical or natural laws. For instance,  stock price prediction. Then, STP techniques require economic justifications to describe BP microstructures and system response. Furthermore,  SPT is powerful when analysing phenomena in the frequency domain \cite{Zhang2017}.

After reviewing relevant work, we found a three-point gap: 1) PBMP requires additional metrics to improve task forecasting \cite{Rama-Maneiro2022, https://doi.org/10.48550/arxiv.1602.04938}. 2) SPT has the potential to provide information to analyse Solow's Paradox redux. However, current strategies only identify cyclical trends \cite{Bogalo2021}. 3)  BPS requires new "what-if" methodologies. Current procedures mainly focus on process reconstruction. Existing approaches employ stochastic methods to reflect effects due to changes in the BP inputs rather than BP structure \cite{Mao2020, Camargo2022}.

\section{Estimating Productivity: Key Definitions}

In this section, we outline key critical definitions used throughout this paper. 

\noindent \textbf{Linear, time-invariant (LTI) system.} An LTI system complies with the requirements of linearity and time-invariance while producing an output signal from any input signal. LTI output of a linear combination of inputs is a linear combination of the individual response to those inputs. LTI system will produce the same result whether we apply an outcome now or in T seconds from now; the only difference will be a T-second delay \cite{alma9927799804401631}. They keep the memory of previous states and can foresee long-term system behaviour. They produce bounded outputs for bounded inputs and cannot generate new frequencies not already present in the intakes \cite{Rao2018}.

\noindent \textbf{Business Process (BP).} A business process is a group of coordinated actions that produce a service or good \cite{Atrill2017}. Business Process Model and Notation (BPMN) is typically used to represent these production chains \cite{Lima2012}. Each task is an abstraction of a production step that is parametrised in terms of production time and skill-set distribution \cite{Kossak2016}. These charts provide information on task ownership, dependencies, information inputs and outputs \cite{Kalnins2004}.
    
\noindent \textbf{Productivity.} Productivity is the relation between the output generated by a BP and the required resources to create this result \cite{Prokopenko1987}. 

\noindent \textbf{Solow’s paradox.}  It is an observed phenomenon that consists of a massive investment in computer systems but not a net gain in productivity.\cite{Denning2021, Brynjolfsson1996, Brynjolfsson1996B}. 

\noindent \textbf{Automation.} "It is the implementation of processes by automatic means" \cite{alma992977422113601631}. In other words, the development and use of technology for production and service delivery monitoring and control.
     
 \section{Methodology}
\begin{figure}[htbp]
            \centering 
                {\includegraphics[scale=0.6]{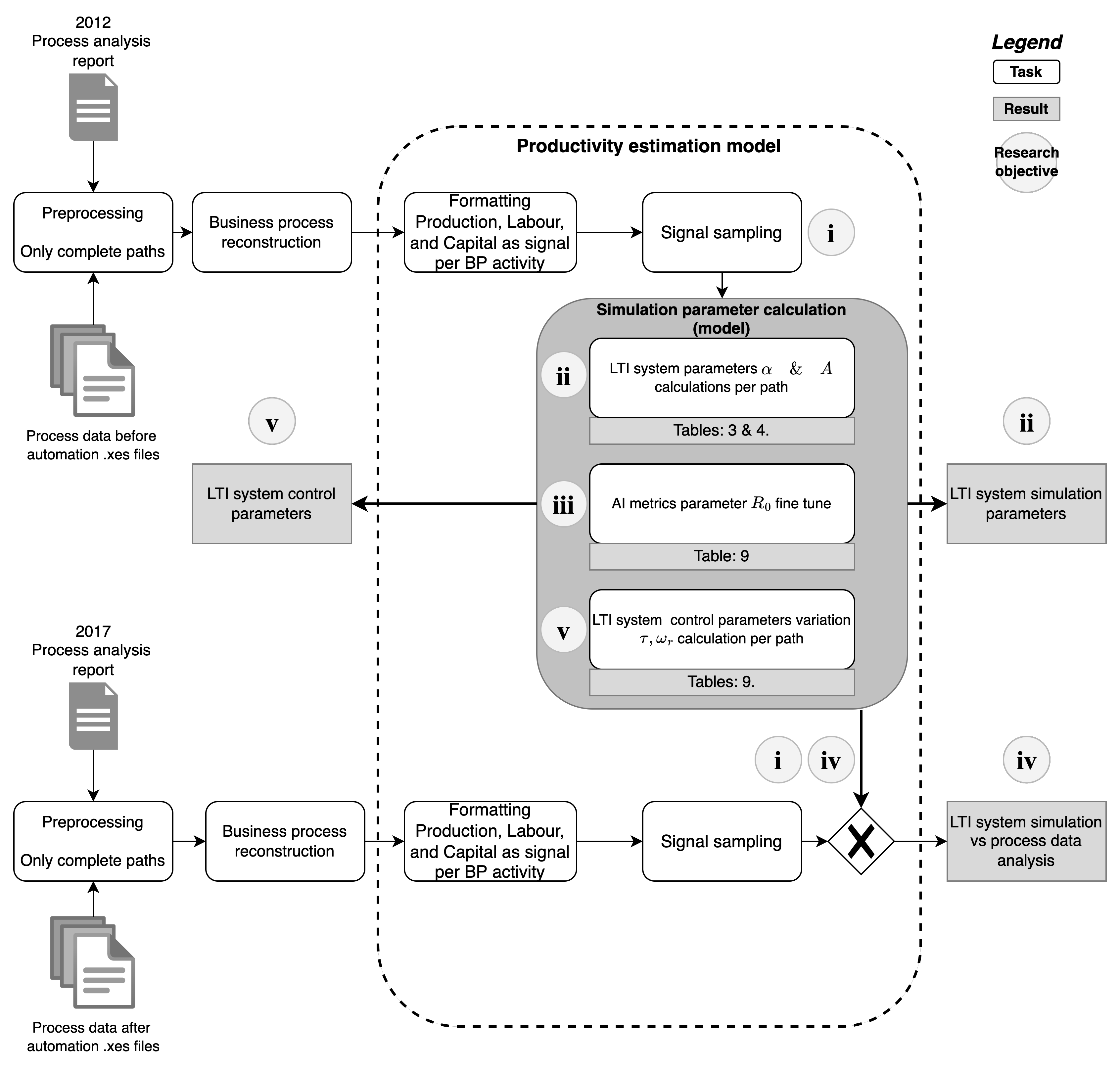}} 
            \caption{Experiment modelling diagram.}
            \label{fig:EXPSTUP} 
        \end{figure}

We propose a novel simulation framework of a BP in the frequency domain to provide a framework for estimating productivity gains. To achieve that, we created the experiment shown in figure \ref{fig:EXPSTUP}.

We examined two datasets from the Business Process Intelligence (BPI) Challenge. They include information on the same process in two different periods, 2012 and 2017 \cite{vanDongen2012, vanDongen2017}. These documents record the pre- and post-state of the automation integration into the process. The datasets also included process description reports. These documents offered details on how labour spread throughout the process.

We considered productivity as a frequency. Given the productivity definition, we have the following:
            \begin{equation}
                P = \frac{Y}{T} [widgets/hr].
                \label{eq:eq1}
            \end{equation}
Where production/output, $Y$, is the estimated volume of products/services generated by a firm, and $T$ is the labour input, the number of worked hours, to produce these outputs \cite{Macdonald2000}. 

A company is a collection of BPs dedicated to delivering a service or manufacturing a product \cite{Atrill2017}. We employed workflows to represent BPs and provide details on task ownership, dependencies, inputs, and outputs. Instead of treating the entire BP as an LTI system, we downsized the analysis and dealt with each BP task as a separate LTI system. Therefore, we can analyse changes in BP's productivity as the shift in BP's task frequency.

Consequently, the characteristics of production as a signal would be as follows, figure \ref{fig:PRDWAVE}:
         \begin{figure}[htbp]
            \centering 
                {\includegraphics[scale=0.6]{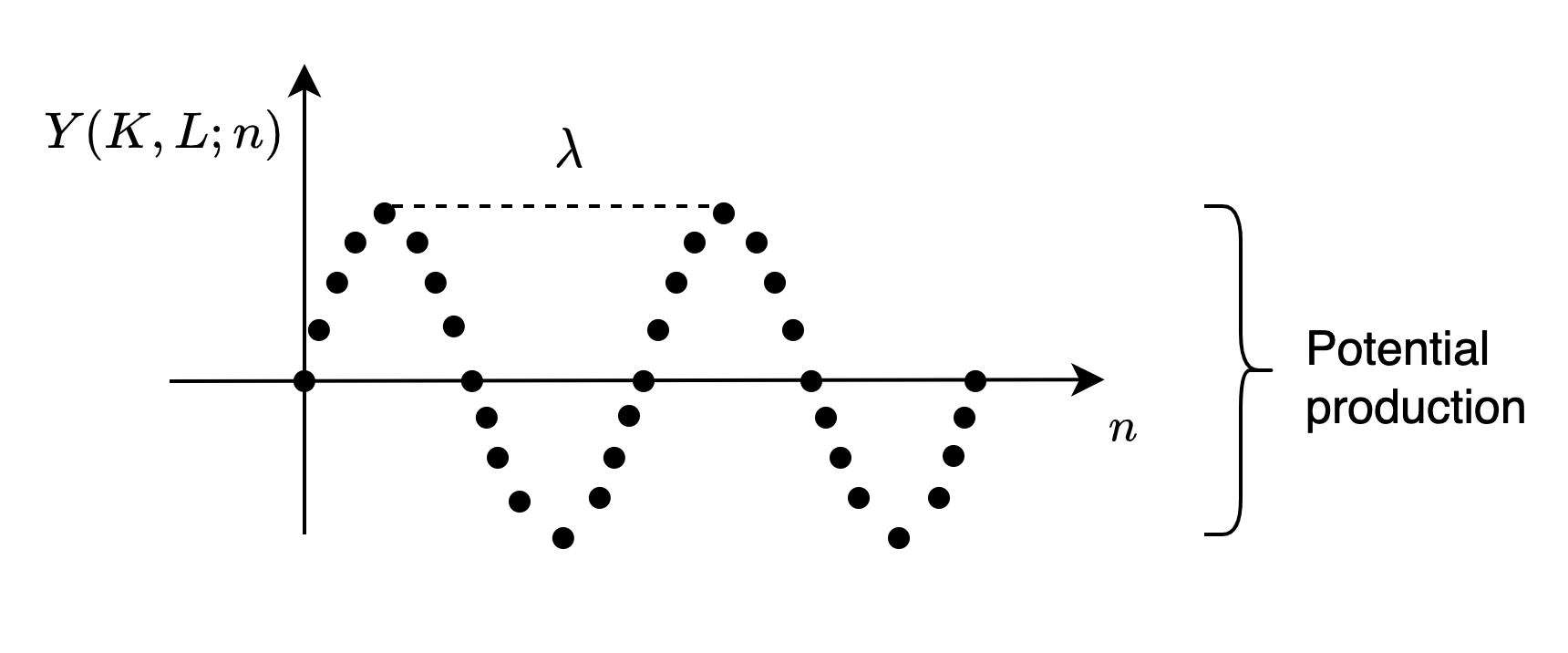}} 
                \caption{Production as a wave in the discrete domain.}
                \label{fig:PRDWAVE} 
         \end{figure}
Where $\lambda$ is the wavelength and the wave frequency,$f$, is given in $[widgets/n]$; $n$ represents the period. 

Our model considers production and its inputs as signals. This approach allows a frequency domain analysis of productivity \cite{Chakravorty2018}.  

\begin{figure}[htbp]
                    \centering 
                        {\includegraphics[scale=0.35]{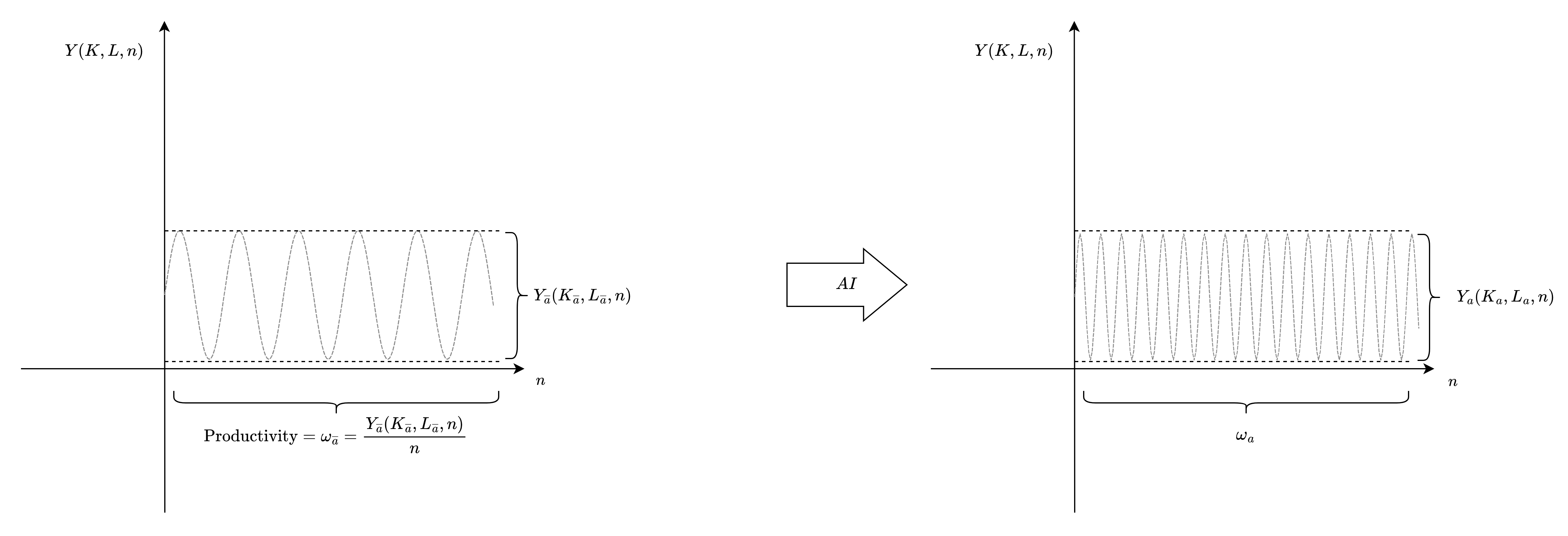}} 
                        \caption{Productivity multiplier}
                        \label{fig:PRDMUL} 
               \end{figure}

We identified three different types of LTI systems for BP-task production: i) initial task, ii) Non-automated task, and iii) Automated task. We considered automated activities as frequency multipliers. See figure \ref{fig:PRDMUL}.

The data from the process logs was regarded as a continuous signal. However, productivity is a digital phenomenon. So, to turn it into a discrete one, we sampled it. To determine the sampling frequency, we employed the Nyquist criterion\cite{sklar2001digital}: 
                    \begin{equation}
                        T_s \leq \frac{total\;labour\;hours}{2\times total\;cases} 
                        \label{eq:TSCALC}
                    \end{equation}
Where $T_s$ is the sampling time/period.

With this background, we applied SPT to identify productivity differences in the form of frequency variations. We produced a report on the  BIP challenge logs. Our model proposes a novel method to analyse productivity variations with SPT. We can scale up or down our analysis by employing the LTI system concept.

\section{Proposed model}
\subsection{Model core}
Our model considers the BP as an LTI system. Therefore, BP production is the sum of each task outcome's contribution. Each BP job is also an LTI itself. Thus, production goes through all the BP activities. In this manner, it is possible to analyse a broad number of BPs by configuring individual tasks and connecting them in the way BPML diagrams attach them.

We inject labour and capital per BP task. Each activity will modify the production and, in some cases, productivity. Changes in production frequency will indicate an increase or decrease in productivity.

We considered the following BP-task inputs: 1) task labour, 2) task capital, 3) and previous task production.

To calculate current task production, we employed Cobb-Douglas equations:
                \begin{equation}
                     Y(t)=A(t)L(t)^{\beta}K(t)^{\alpha}
                     \label{eq:CobbDouglas}
                \end{equation}
Where $K(t)$ represents capital and $L(t)$ is labour. $A$, $\alpha$ and $\beta$ are usually constants\cite{Heathfield1971}. To linearise equation \ref{eq:CobbDouglas}, we apply natural logarithms to both sides of the equality. Moreover, $\alpha$ and $\beta$ are complementary: $\alpha + \beta = 1$. Representing $\beta$  in terms of $\alpha$, we have the following:
                \begin{align}
                      \ln{Y(t)} &= \ln{\{A(t) L(t)^{(1-\alpha)} K(t)^\alpha \}} \notag \\
                       \ln{Y(t)} &= \ln{A(t)} + (1-\alpha) \ln{L(t)} + \alpha \ln{K(t)}
                    \label{eq:LGPROD}
                \end{align}

We can determine $A$ and $alpha$ values using linear regression algorithms with the expression \ref{eq:LGPROD}. For this calculation, we might use production log information.

Equation \ref{eq:CobbDouglas} only relates the first two inputs. For the last intake, we used the multi-input production model \cite{Aghion2017}:
                \begin{equation}
                      Y = QX_1^{\lambda_1}X_2^{\lambda_2}\cdot\textit{}\dots \textit{}\cdot X_n^{\lambda_n} \text{  where  } \sum_{i=1}^{n} \lambda_i = 1 .
                      \label{eq:MULTFPRODC1}
                \end{equation}

Where $Q$ is an efficiency parameter, $Xs$ are inputs and $\lambda s$ are elasticity parameters. 

We can consider that equation \ref{eq:MULTFPRODC1} has only two inputs: the production of the current activity, $Y_a$, and the production of the previous task, which reflects the participation of earlier inputs, $Y_i$. Therefore: 
                    \begin{align}
                        Y_o &= Q Y_a^{\lambda_1} Y_i^{\lambda_2} \text{  where  } \sum_{k=1}^{2} \lambda_k = 1 .  
                        \label{eq:MULTFPRODCF}
                    \end{align}

The zero initial conditions are one of the fundamental characteristics of LTI systems \cite{alma9931362004401631}. If $Y_o$ = 1, in equation \ref{eq:MULTFPRODCF}, $Y_a$ and $Y_i$ should be 1 because it is the only possible value that maintains the identity. Therefore, $Q$ should also be 1. As a result, $Q=1$ or $\ln{Q}=0$. 
Furthermore, our model deduces that both inputs equally contribute to $Y_o$. Consequently, $\lambda_1 = \lambda_2$. Thus, $\lambda_1=\lambda_2=\frac{1}{2}$. The resultant expression is the geometric average of the current activity's production, $Y_a(t)$, and the previous task's, $Y_i(t)$, production.

Hence, we have: 
           \begin{align}
                Y_o(t) &=  Y_a(t)^{\frac{1}{2}} Y_i(t)^{\frac{1}{2}} \notag  \\
                Y_o(t) &= \sqrt{ A_a(t) L_a(t)^{(1-\alpha)}K_a(t)^{\alpha} Y_i(t)}
                \label{eq:NLDEF}
           \end{align}

Applying logarithms to the equation \ref{eq:NLDEF} and re-writing it, e.g. $\ln{X(t)} \xrightarrow{} x[t]$, we have:
           \begin{align}
                y_o[t] &= \frac{1}{2} \bigg [ a_a[t] + (1-\alpha) l_a[t] + \alpha \: k_a[t] + y_i[t] \bigg  ]
                \label{eq:LGPRODRES}
           \end{align}

Equation \ref{eq:LGPRODRES} represents our model in continuous time. Equation \ref{eq:PRODDISD} shows the results of sampling $Y_o$ with the sample period $T_s$:
                    \begin{align}
                        y_o[nT] &= \frac{1}{2} \bigg [  a_a[nT] + (1-\alpha) l_a[nT] + \alpha k_a[nT] + y_i[nT]   \bigg ]   \notag \\  \notag \\ 
                        y_o[n] &= \frac{1}{2} \bigg [  a_a[n] + (1-\alpha) l_a[n] + \alpha k_a[n] + y_i[n] \bigg ] 
                        \label{eq:PRODDISD}
                    \end{align}

Equation \ref{eq:PRODDISD} is our general BP task production model.  Nevertheless, it has a different shape for every task type. The following expressions describe the behaviour of every chore type:
            \begin{itemize}
                \item  Initial task model:
                    \begin{equation}
                        y_o[n] = 
                        \begin{cases}
                                  a_a[n] + (1-\alpha) l_a[n] + \alpha k_a[n], & \text{if  } L_a[n] > 0 \text{ and } K_a[n] > 0\\
                                \\
                                0, & \text{otherwise. }
                        \end{cases}
                        \label{eq:PRODINTSK}
                    \end{equation}
                \item Non-automated task model:
                    \begin{equation}
                        y_o[n] = 
                        \begin{cases}
                                \frac{1}{2} \bigg [  a_a[n] + (1-\alpha) l_a[n] + \alpha k_a[n] + y_i[n] \bigg ], & \text{if  } L_a[n] > 0 \text{ , }  K_a[n] > 0 \text{ and } Y_i[n] > 0\\
                                \\
                                y_i[n], & \text{otherwise. }
                                            
                        \end{cases}
                        \label{eq:PRODNATSK}
                    \end{equation}
                \item As for $a[n]$:
                    \begin{equation}
                        a[n] = 
                        \begin{cases}
                                C, & \text{if  } L_a[n] > 0 \text{ and }  K_a[n] > 0\\
                                \\
                                0, & \text{otherwise. }
                        \end{cases}
                    \end{equation}
              Where $C$ is a constant value.
                    
            \end{itemize}

\subsection{The automated task}
We modelled each BP task as an LTI system. LTIs do not alter the frequency of the input signal. We used the concept of Amplitude Modulation, AM, to change this behaviour \cite{LianjunAn}. AM is a type of broadcasting; it moves a signal's message frequency up toward its carrier frequency \cite{alma9927799804401631}. AM definition is the multiplication of the message and the carrier signals. We employed this frequency shift to boost BP task productivity. 

In our case, $Y_o[n]$ is the message, $c[n]$ is the carrier, and $y[n]$ is the improved production signal. We employed the following carrier function $c[n]$:
            \begin{align}
                c[n] = \cos^2{\omega_c} = \frac{1}{2} (1 + \cos{2 \omega_c n})
            \end{align}
Where $\omega_c = 2  \pi(1+\kappa)f_m$. $f_m$ is the average frequency of production signal and $\kappa$ is the increase in productivity due to automation.

Then, our automated task model is as follows:
                    \begin{equation}
                        y_o[n] = 
                        \begin{cases}
                                \frac{1}{2} \bigg [  a_a[n] + (1-\alpha) l_a[n] + \alpha k_a[n] + y_i[n] \bigg ] \bigg [ \frac{1}{2} (1 + \cos{2 \omega_c n}) \bigg ], & \text{if  } L_a[n] > 0 \text{ , }  K_a[n] > 0 \text{ and } Y_i[n] > 0\\
                                \\
                                y_i[n], & \text{otherwise. }         
                        \end{cases}
                        \label{eq:PRODAUTSK}
                    \end{equation}
Where $a_a[n]$:
                    \begin{equation}
                        a_a[n] = 
                        \begin{cases}
                                C, & \text{if  } L_a[n] > 0 \text{ , }  K_a[n] > 0 \text{ and } Y_i[n] > 0\\
                                \\
                                0, & \text{otherwise. }
                        \end{cases}
                    \end{equation}

It is possible to introduce digital control techniques because we are working in the frequency domain. To examine how metrics affect the model, we include a negative feedback loop. As a result, we can evaluate productivity patterns brought on by changes in task output. We suggest using accuracy as a system metric for automation. See equation \ref{eq:PSAUSMET}.
                    \begin{equation}
                        A_{cc}[n] := R_o (1 - e^{-\frac{1}{\tau} n}) 
                        \label{eq:PSAUSMET}
                    \end{equation}
                    
$\tau$ allows us to simulate accuracy behaviour over time, and $R_o$ is the targeted accuracy. It has the following definition:
                    \begin{equation}
                        R_o = \frac{True_{positive_o} + True_{negative_o}}{True_{positive_o} + True_{negative_o}+False_{positive_o} + False_{negative_o}}
                    \end{equation}

If we apply the $Z$ transform to equations \ref{eq:PRODINTSK}, \ref{eq:PRODNATSK} and \ref{eq:PRODAUTSK}, we have the following:
                    \begin{equation}
                        Y_o[z] =
                                \begin{cases}
                                     \frac{1}{2} \bigg[ A_a[z] + ( 1 - \alpha) L_a[z] + \alpha K_a[z] + Y_i[z]   \bigg ], & \text{if  } L_a[z] > 0 \text{ and }  K_a[z] > 0 \text{ and } Y_i[z] > 0\\ 
                                     Y_i[z], & \text{ otherwise}.
                                \end{cases}
                        \label{eq:PSNAUZT}
                    \end{equation}
                        
                    \begin{equation}
                         Y_o[z] =
                                \begin{cases}
                                      \bigg (
                                         \frac{e^{\frac{1}{\tau}}z^2 -(e^{\frac{1}{\tau}} + 1)z +1}{e^{\frac{1}{\tau}}z^2 + ((e^{\frac{1}{\tau}} - 1)R_o - (e^{\frac{1}{\tau}} + 1))z + 1} (\frac{1}{2}) \bigg[ A_a[z] + ( 1 - \alpha) L_a[z] + \alpha K_a[z] + Y_i[z]   \bigg ] \bigg ) * C[z] &, \\
                                         \qquad \qquad \qquad \qquad \qquad \qquad \qquad \qquad \qquad \qquad \text{if  }  L_a[z] > 0  \text{ and } K_a[z] > 0 \text{ and } Y_i[z] > 0\\
                                          Y_i[z] &, \text{ otherwise}. 
                                \end{cases}
                        \label{eq:PSAUZT}
                    \end{equation}
                    
\begin{figure}[!htb]
                    \minipage{0.33\textwidth}
                        \includegraphics[width=\linewidth]{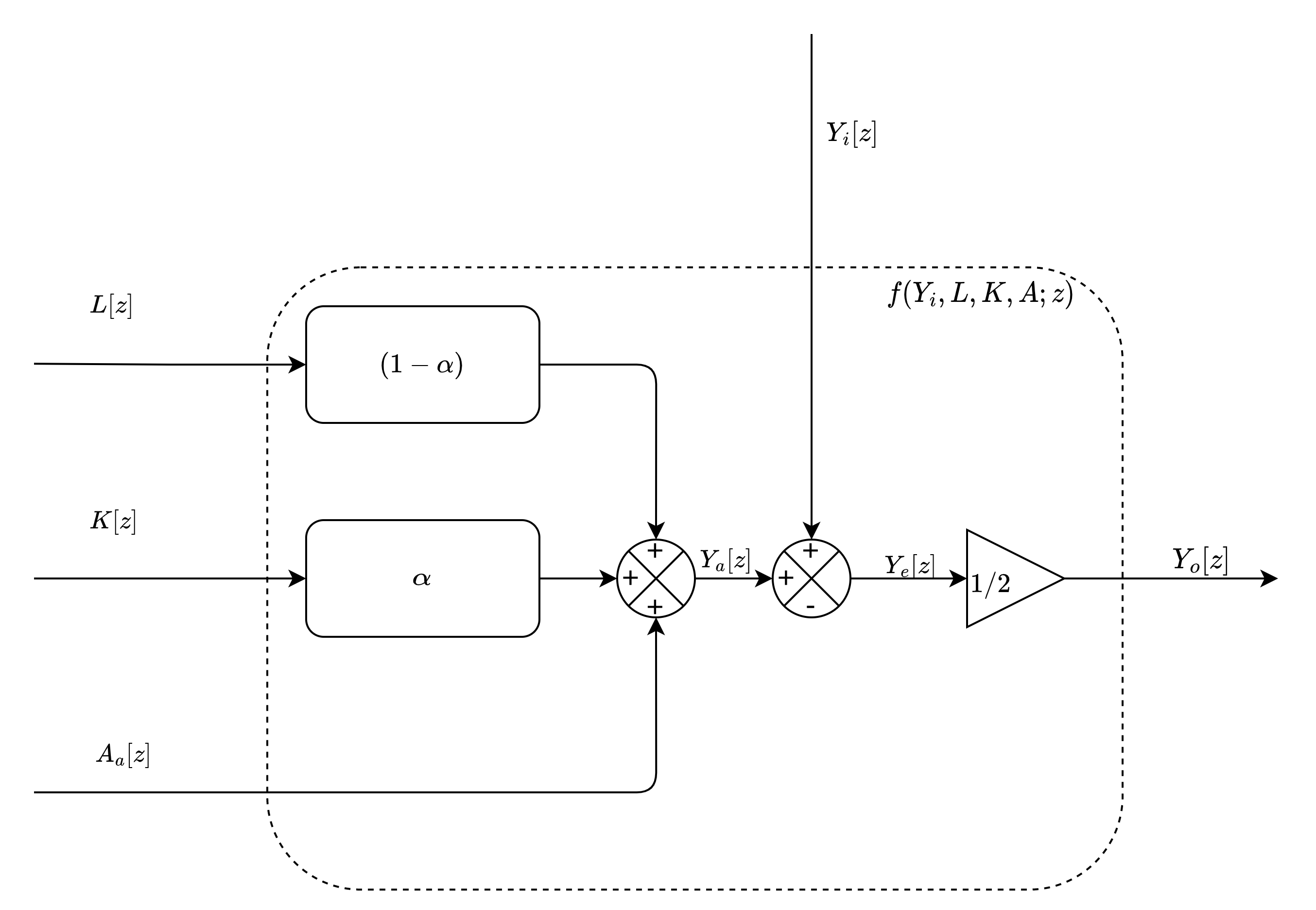}
                        \caption{Non-automated task.}
                        \label{fig:PRDBDMD}
                    \endminipage\hfill
                    \minipage{0.55\textwidth}
                        \includegraphics[width=\linewidth]{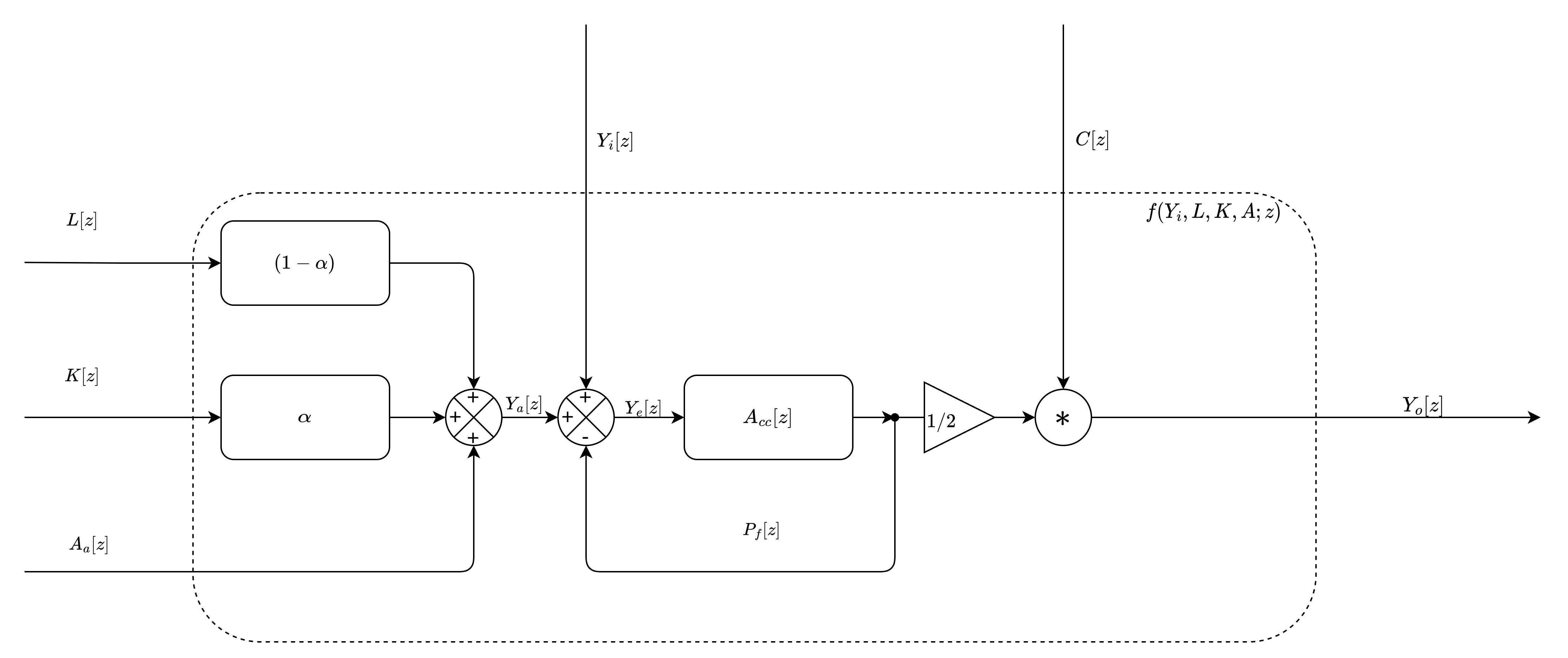}
                        \caption{Automated task.}
                        \label{fig:PRDBDAD}
                    \endminipage\hfill
                \end{figure}

Where $*$ is the convolution on $Z$ domain operation. See figures \ref{fig:PRDBDMD}, and \ref{fig:PRDBDAD}.

\section{Results}
    \subsection{Productivity performance}
\begin{figure}[!htbp]
           \centering 
               {\includegraphics[scale=0.3]{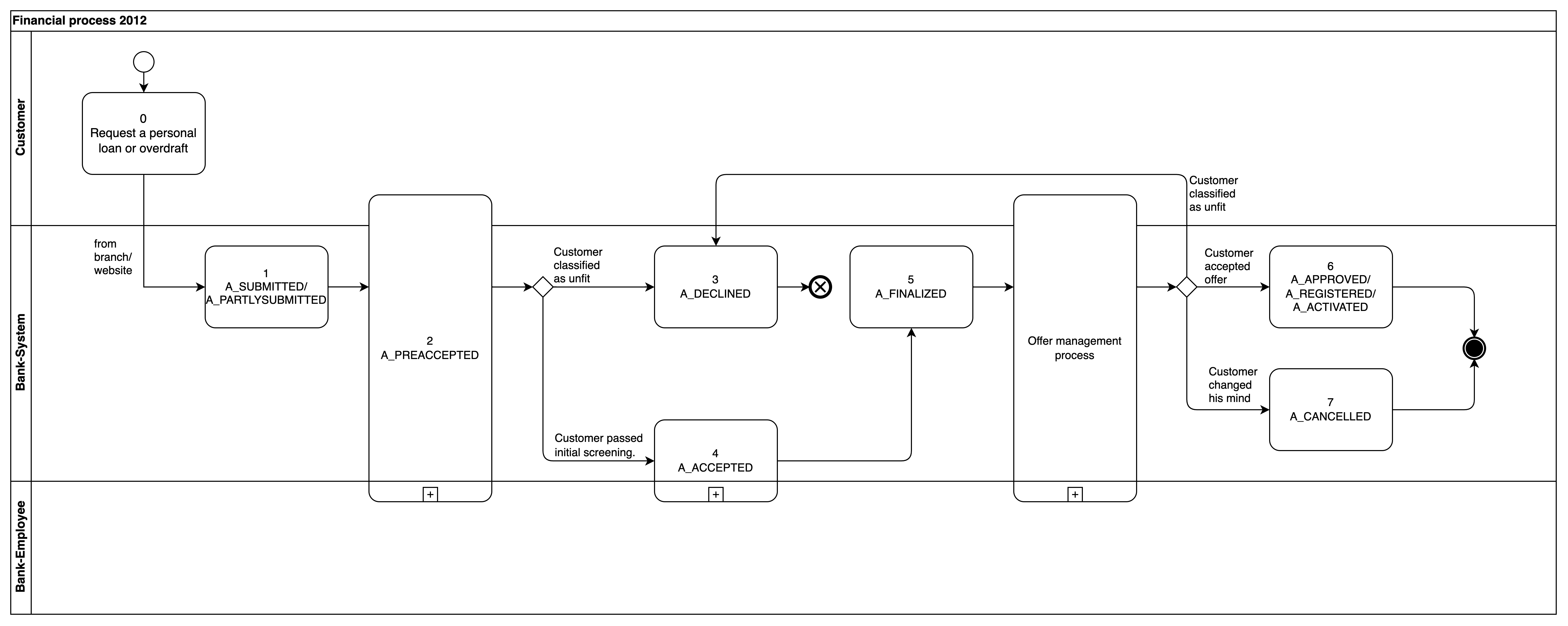}} 
            \caption{Business process 2012.}
            \label{fig:BP2012} 
       \end{figure}

\begin{figure}[!htbp]
            \centering 
                {\includegraphics[scale=0.06]{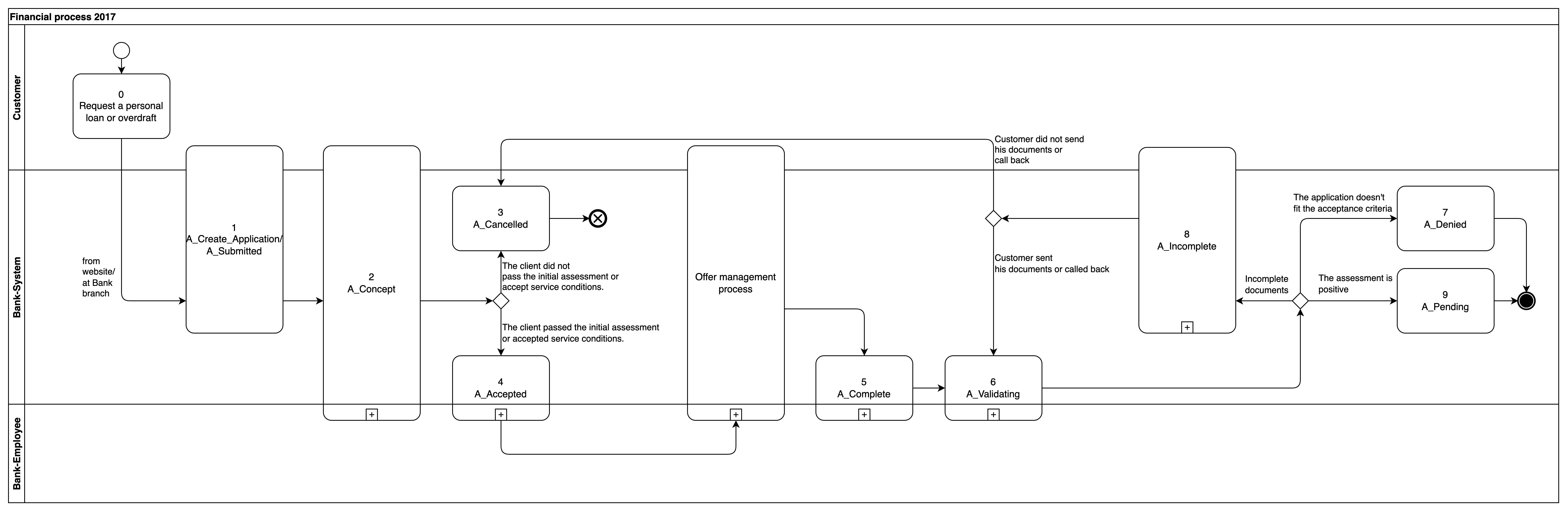}} 
             \caption{Business process 2017.}
             \label{fig:BP2017} 
       \end{figure}

We recreated the BPs found in the logs from 2012 and 2017 \cite{vanDongen2012, vanDongen2017}. Figures \ref{fig:BP2012}, and  \ref{fig:BP2017}. We determined each process's "main paths." A series of actions that move from an initial state to a final state is referred to as a "main path" \cite{Kalnins2004}.

We found the paths in the table \ref{tab:MP1217}. Path name contains the initial and final state. We encountered that routes from both years are equivalent. In 2017's log, all trails include an automated task and the state name also changed.

Paths enclose the following cases: loan documentation did not accomplish the institution's requirements, is approved, and is rejected, respectively. Either banks or customers can forsake a loan.

       \begin{table}[!htbp]
            \begin{tabular}{|c|c|c|}
                \hline
                    Path name & 2012 & 2017 \\ 
                    \hline 
                    A & $A\_SUBMITTED \rightarrow A\_CANCELLED$   & $A\_Create \quad Application \rightarrow  A\_Denied$ \\ \hline 
                    B & $A\_SUBMITTED \rightarrow A\_REGISTERED$   &
                    $A\_Create \quad Application \rightarrow A\_Pending$ \\ \hline 
                    C & $A\_SUBMITTED \rightarrow A\_DECLINED$   &  
                    $A\_Create \quad Application \rightarrow A\_Cancelled$ \\ \hline
            \end{tabular}
            \caption{Main paths 2012 and 2017}
            \label{tab:MP1217}
       \end{table}

We considered paths as cumulative. Thus, production at time $n$ is the
result of all the courses ``alive” in the system.    

We got the following results when we apply equation \ref{eq:TSCALC} to the logs:
        \begin{table}[!htbp]
                \centering
                \begin{tabular}{|c|c|c|}
                     \hline
                          & 2012 & 2017 \\ 
                     \hline 
                          $T_s$ & 5 [minutes]   &  5 [minutes]  \\ \hline 
                \end{tabular}
                \caption{Sampling time for logs 2012 and 2017}
                \label{tab:TS1217}
        \end{table}

From this point forward, to make the information analysis easier: 1 [$Hertz/Hz$] = 1 / ${T_s}$ = 1/(5 [$minutes$]). 

We calculated $A_a(t)$ and $\alpha$ for each path as the initial step. To estimate Cobb-Douglas parameters, we used linear regression and equation \ref{eq:LGPROD}. An example of path A for the 2012 and 2017 processes are shown in tables \ref{tab:L2012PATHA} and \ref{tab:L2017PATHA}.

                \begin{table}[!htb]
                        \begin{minipage}{.5\linewidth}
                            \centering
                                    \begin{tabular}{|l|c|c|c|}
                                    \hline
                                    Activity & $A_a(t)$ & $\alpha$ & $f_m [Hz]$ \\
                                    \hline
                                    A\_SUBMITTED   & N/A   & N/A            & 0 \\ \hline
                                    A\_PREACCEPTED & -0.18 & 0.02 & 0.14 \\ \hline
                                    A\_ACCEPTED    & -1.11 & 0.11 & 0.07 \\ \hline
                                    A\_FINALIZED   & -0.53 & 0.06 & 0.14 \\ \hline
                                    A\_CANCELLED   & -3.29 & 0.17 & \num{1.29e-05} \\ \hline
                                    \end{tabular}
                                    \caption{Log 2012 - Path A}
                                    \label{tab:L2012PATHA}
                        \end{minipage}%
                        \begin{minipage}{.5\linewidth}
                            \centering
                                \begin{tabular}{|l|c|c|c|}
                                    \hline
                                         Activity & $A_a(t)$ & $\alpha$ & $f_m [Hz]$ \\
                                        \hline
                                        A\_Submitted   & N/A   & N/A            & 0 \\ \hline
                                        A\_Concept & -0.07 & 0.01 & 0.13 \\ \hline
                                        A\_Accepted    & -1.62 & \num{2.12e-03} & \num{1.67e-04} \\ \hline
                                        A\_Complete   & -1.38 & 0.02 & \num{2.96e-03} \\ \hline
                                        A\_Validating   & -3.42 & 0.18 & \num{6.67e-06} \\ \hline
                                        A\_Denied   & -4.94 & 0.42 & \num{4.01e-05} \\ \hline
                                    \end{tabular}
                                    \caption{Log 2017 - Path A}      \label{tab:L2017PATHA}
                        \end{minipage} 
                \end{table}

For all the signals, we used medium frequency, $f_m$, as a reference. Table \ref{tab:AVGFRP1217} shows productivity, $f_m$, per path:
        \begin{table}[!htb]
            \centering
                \begin{tabular}{|c|c|c|c|}
                    \hline
                    Path name & $f_{m_{2012}} [Hz]$ & $f_{m_{2017}} [Hz]$ & $f_{wom_{2017}} [Hz]$\\ 
                    \hline 
                    A & \num{1.39e-05} & \num{3.84e-06} & \num{1.01e-05} \\ \hline 
                    B & \num{5.94e-06} & \num{5.62e-06} & \num{7.28e-06} \\ \hline 
                    C & \num{1.35e-05} & \textbf{\num{3.23e-06}} & \num{5.9e-05} \\ \hline
                \end{tabular}
                \caption{Average frequencies per path}
                \label{tab:AVGFRP1217}
        \end{table}

We can see that the frequencies recorded in 2017, $f_{m_{2017}}$ are lower than those from 2012, $f_{m_{2012}}$. This result is consistent with previous analyses of productivity \cite{Jacobo-Romero2021}. The highlighted value is the lowest frequency.

\begin{figure}[!htbp]
                \centering 
                    {\includegraphics[scale=0.396]{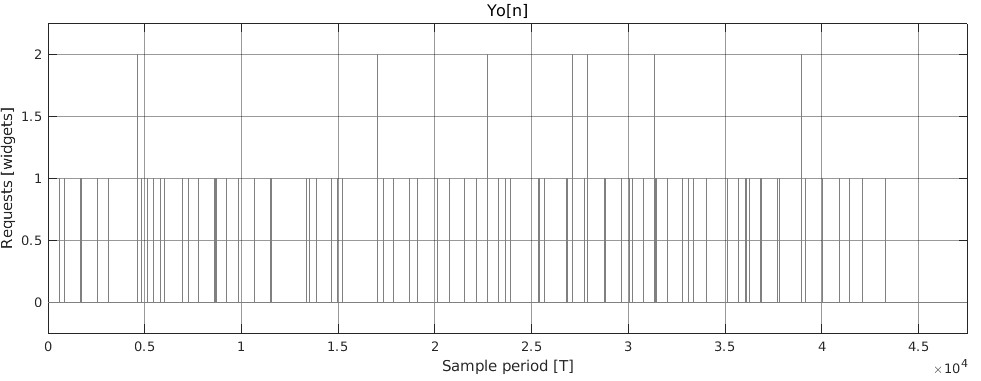}} 
                \caption{``A\_PREACCEPTED'' - Business process 2012. Low frequency.}
                \label{fig:ATTRES1} 
            \end{figure}

\begin{figure}[!htbp]
                \centering 
                    {\includegraphics[scale=0.34]{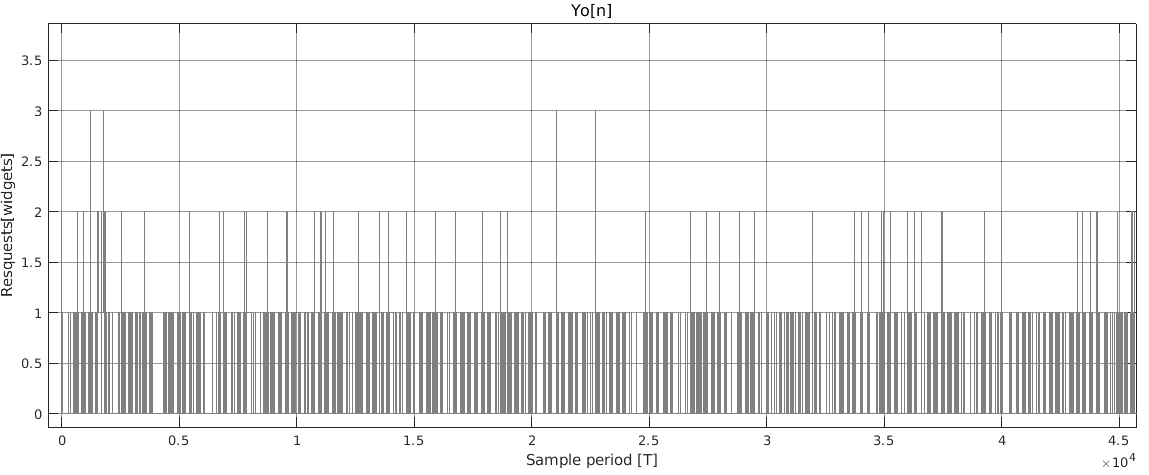}} 
                \caption{``A\_Concept'' - Business process 2017. High frequency.}
                \label{fig:ATTRES2} 
            \end{figure}

The activity ``A\_PREACCEPTED'' was automated and renamed to ``A\_Concept''. The new task is an AI system that executes profiling, classification and suggestion tasks \cite{Lima2012, Povalyaeva2017BPIC2}. Figure \ref{fig:ATTRES1} and \ref{fig:ATTRES2} show production signals from original files.
   
We can see that the introduction of automation increased the frequency of production. Value variations reveal this transformation. We can observe that ``A\_Concept'' changed its frequency. Table \ref{tab:AVGFRT1217} shows these changes per path.

               \begin{table}[!htb]
                \centering
                    \begin{tabular}{|c|c|c|c|}
                     \hline
                        Path name & $f_{m_{2012}} [Hz]$  & $f_{m_{2017}} [Hz]$ & $f_{wom_{2017}} [Hz]$\\
                         & ``A\_PREACCEPTED'' &  ``A\_Concept'' & ``A\_Concept'' \\
                        \hline 
                        A & \num{1.35e-01} &  \num{1.20e-01} & \num{1.33e-01}  \\ \hline 
                        B & \num{4.64e-02} & \num{1.11e-01} & \num{1.22e-01}  \\ \hline 
                        C & \num{4.57e-02} & \num{9.85e-02} & \num{1.08e-01}  \\ \hline
                    \end{tabular}
                    \caption{Intervened task-average frequencies per path}
                    \label{tab:AVGFRT1217}
                \end{table}

Automation not only simplifies tasks but also creates new ones. New duties can be automatic or manual. Manual tasks introduced a time overhead in the process \cite{Povalyaeva2017BPIC2}. The medium frequency, $f_{wom}$, was recalculated after we had filtered out manual tasks. As compared to earlier findings, we can see that productivity variations have improved in most of the cases. 

We also noticed alterations in the fundamental frequencies per path. The behaviour in table \ref{tab:AVGFHL1217} is similar to that in table \ref{tab:AVGFRP1217}.

            \begin{table}[!htb]
               \centering
                   \begin{tabular}{|c|c|c|c|}
                    \hline
                        Path name & $f_{0_{2012}} [Hz]$  & $f_{0_{2017}} [Hz]$ & $f_{wo0_{2017}} [Hz]$\\
                         & ``A\_PREACCEPTED'' &  ``A\_Concept'' & ``A\_Concept'' \\
                        \hline 
                        A & \num{3.45e-03} & \num{4.91e-04} & \num{4.91e-04} \\ \hline 
                        B & \num{1.18e-03} & \num{1.47e-03} & \num{3.49e-03} \\ \hline 
                        C & \num{6.31e-05} & \num{2.90e-04} & \num{2.90e-04} \\ \hline
                  \end{tabular}
                  \caption{Fundamental frequencies per path.}
                 \label{tab:AVGFHL1217}
            \end{table}

The same behaviours as $f_m$ are also displayed by $f_0$. Therefore, empirical evidence supports our claim that automation increases production frequency. 

Based on $f_m$, we calculated the improvement rate.  We called this proportion $\kappa$ \cite{Jacobo-Romero2021}. Table \ref{tab:AVGFRT1217} shows $\kappa$ values:
            \begin{table}[!htb]
               \centering
                    \begin{tabular}{|c|c|c|}
                    \hline
                        Path name & $\kappa_{m_{2017}} [\%]$ & $\kappa_{wom_{2017}} [\%]$\\
                               &  ``A\_Concept'' & ``A\_Concept'' \\
                        \hline 
                        A & -11.52 &  -1.82  \\ \hline 
                        B & 139.17 & 162.00  \\ \hline 
                        C & 115.84 & 137.40  \\ \hline
                    \end{tabular}
                    \caption{Kappa values path per automated task.}
                    \label{tab:KAPPAV1217}
            \end{table}     
            
\subsection{Automated-task stability analysis}
Our analysis included the effect of metrics on automated tasks. Typically, metrics affect production. They limit task output depending on the agreed measure. 

Equation \ref{eq:PSAUSMET} is a function that we proposed to mimic the adoption of a general metric. As shown in figure \ref{fig:SACCT5R090}, we integrated our benchmark as a closed-loop system rather than performing this operation in the time domain.

The automated task equation integrates the transfer function of the metric see equation \ref{eq:PSNAUZT}. We applied the superposition principle to analyse the system stability and  got  the following results:

                \begin{figure}[!htb]
                    \minipage{0.32\textwidth}
                        \includegraphics[width=\linewidth]{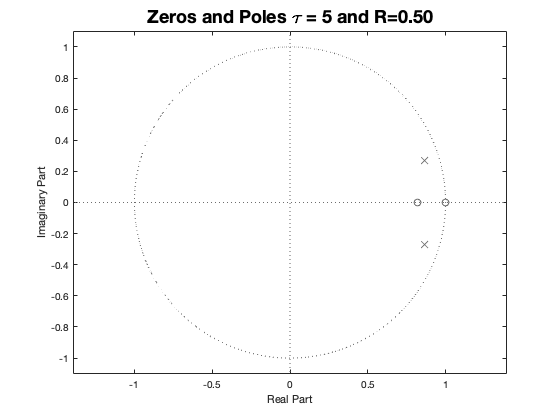}
                        \caption{$\tau = 5$ and \\ $R=0.50$}
                        \label{fig:SACCT5R050}
                    \endminipage\hfill
                    \minipage{0.32\textwidth}
                        \includegraphics[width=\linewidth]{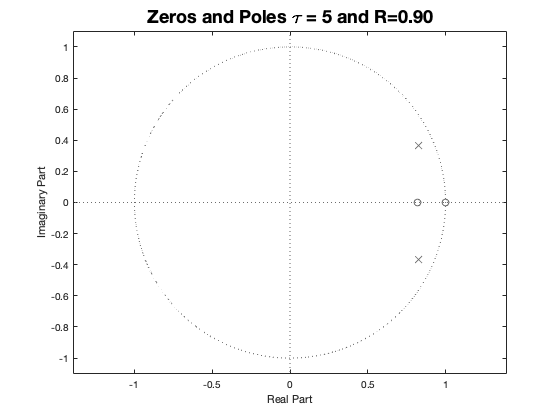}
                        \caption{$\tau = 5$ and \\ $R=0.90$}
                        \label{fig:SACCT5R090}
                    \endminipage\hfill
                    \minipage{0.32\textwidth}%
                        \includegraphics[width=\linewidth]{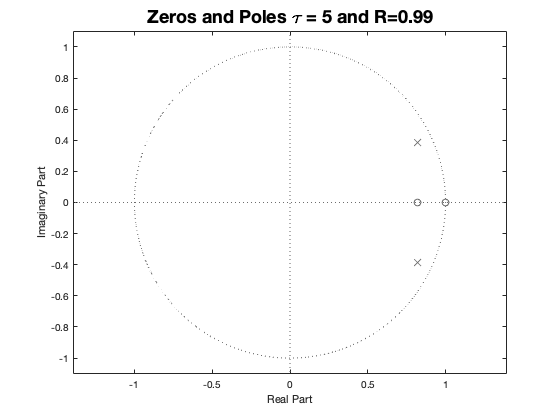}
                        \caption{$\tau = 5$ and \\ $R=0.99$}
                        \label{fig:SACCT5R099}
                    \endminipage
                \end{figure}
                
                \begin{figure}[!htb]
                    \minipage{0.32\textwidth}
                        \includegraphics[width=\linewidth]{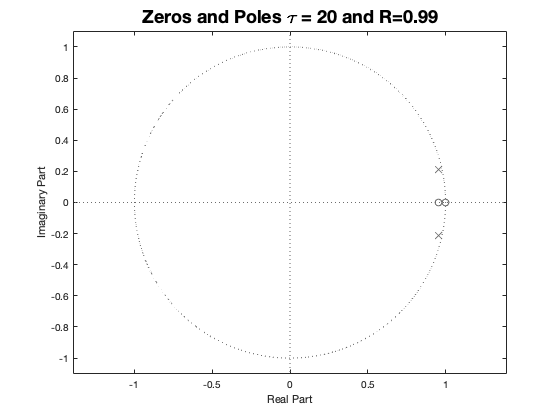}
                        \caption{$\tau = 20$ and \\ $R=0.99$}
                        \label{fig:SACCT20R099}
                    \endminipage\hfill
                    \minipage{0.32\textwidth}
                        \includegraphics[width=\linewidth]{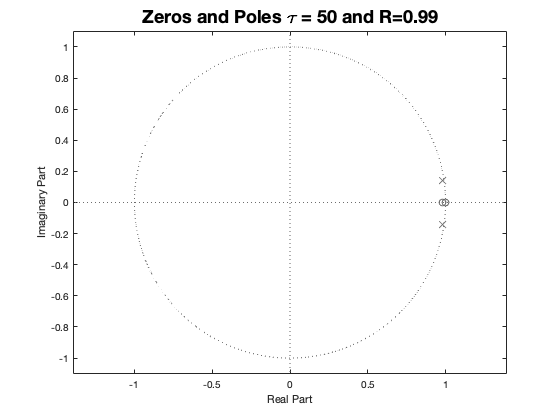}
                        \caption{$\tau = 50$ and \\ $R=0.99$}
                        \label{fig:SACCT50R099}
                    \endminipage\hfill
                    \minipage{0.32\textwidth}%
                        \includegraphics[width=\linewidth]{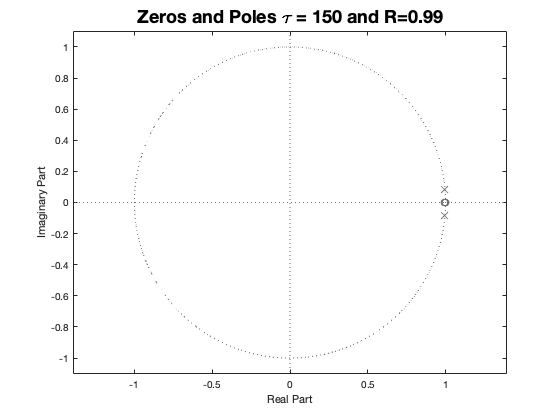}
                        \caption{$\tau = 150$ and \\ $R=0.99$}
                        \label{fig:SACCT150R099}
                    \endminipage
                \end{figure}

Figures \ref{fig:SACCT5R050} to \ref{fig:SACCT150R099} show the stability diagram for different values of $tau$ and $R_o$.  A predictable/stable output should own poles, $x$, within the unitary circle. 

Our tests showed that for $ \tau < 50$, "poles" and "zeros", $o$,  are closer to the circle's centre. On the other hand, "poles" travel to the instability boundaries when $R_o$ values are near 1. 

After several tests, we empirically found that optimum values for  $R_o$ and $\tau$ are $0.99$ and $20$, respectively. These parameters allowed us to determine the resonance frequency of our model. See table \ref{tab:RFACCM}.

                \begin{table}[htbp]
                    \centering
                    \begin{tabular}{|c|c|c|c|}
                        \hline
                        &  $\text{Resonance Frequency} [Hz]$ & $R_0$ & $\tau$\\
                        \hline 
                        $\omega_r$ & 1.03 & 0.99 & 20 \\ \hline 
                    \end{tabular}
                    \caption{Resonance frequency of the accuracy metric. }
                    \label{tab:RFACCM}
                \end{table}

We can't guarantee the predictability and stability of the system if the automated task approaches $\omega_r$. Beyond this frequency, the model's output rapidly decreases. Converting $\omega_r$ to log units \cite{vanDongen2012, vanDongen2017},  we have the following: $1.03 [Hz] = 1.03/5 [requests/min] \approx 12 [requests/hr] $
    
\subsection{Simulation}

We simulated the 2012's process using data from table \ref{tab:KAPPAV1217} in MATLAB\textsuperscript{\tiny\textregistered}. All of our tasks were created in Simulink\textsuperscript{\tiny\textregistered} as blocks. Using this method, we simulated 4 different process iterations. They covered all possible paths.

Each block has several inputs. The initial task receives only Capital and Labour input signals. We parsed, sampled, and stored log information in MS Excel\textsuperscript{\tiny\textregistered} format. Automated and no-automated activity blocks receive production signals from previous blocks to compute the influence of current duty in the BP's response. 

\begin{figure}[!htb]
                    \centering
                        \begin{minipage}{.5\textwidth}
                            \centering
                            \includegraphics[scale=0.12]{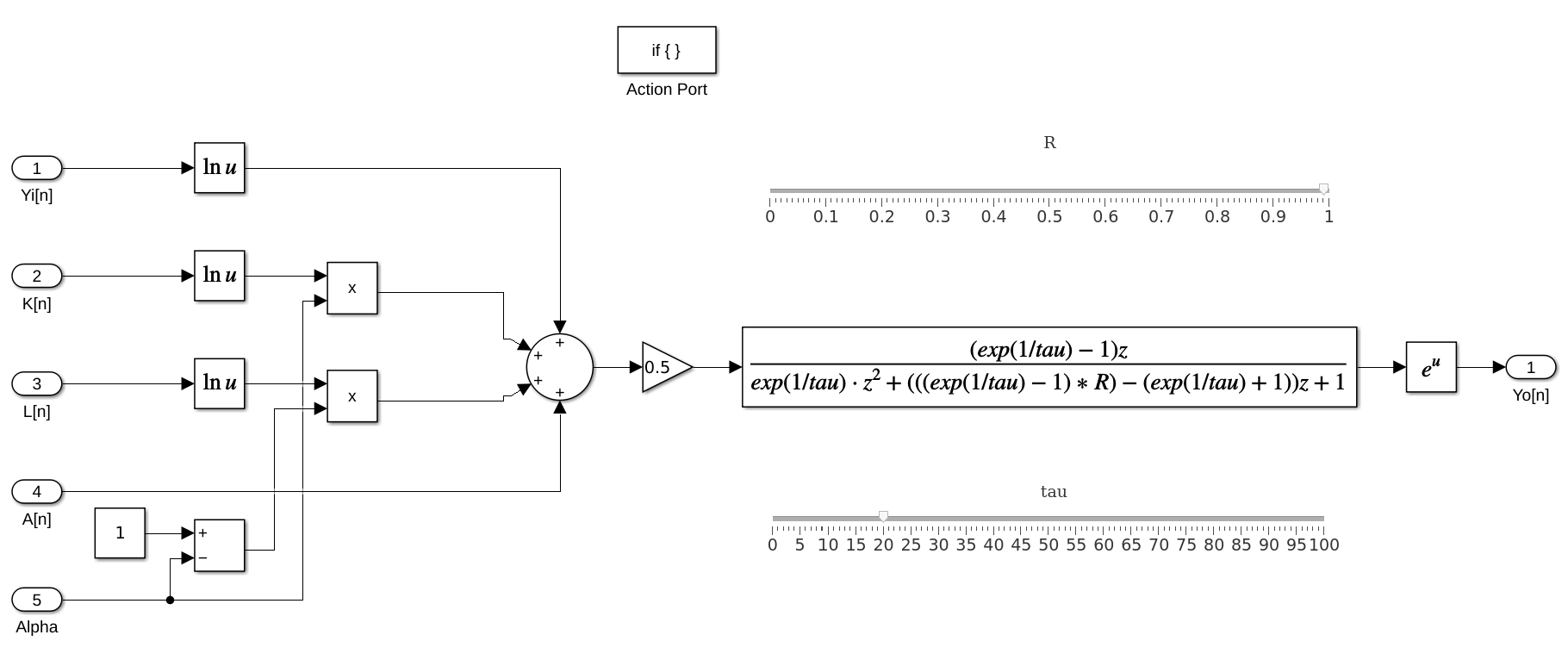}
                             \caption{Automated Task Model - I.}
                             \label{fig:AUTTASKI}
                         \end{minipage}%
                         \begin{minipage}{.6\textwidth}
                             \centering
                             \includegraphics[scale=0.12]{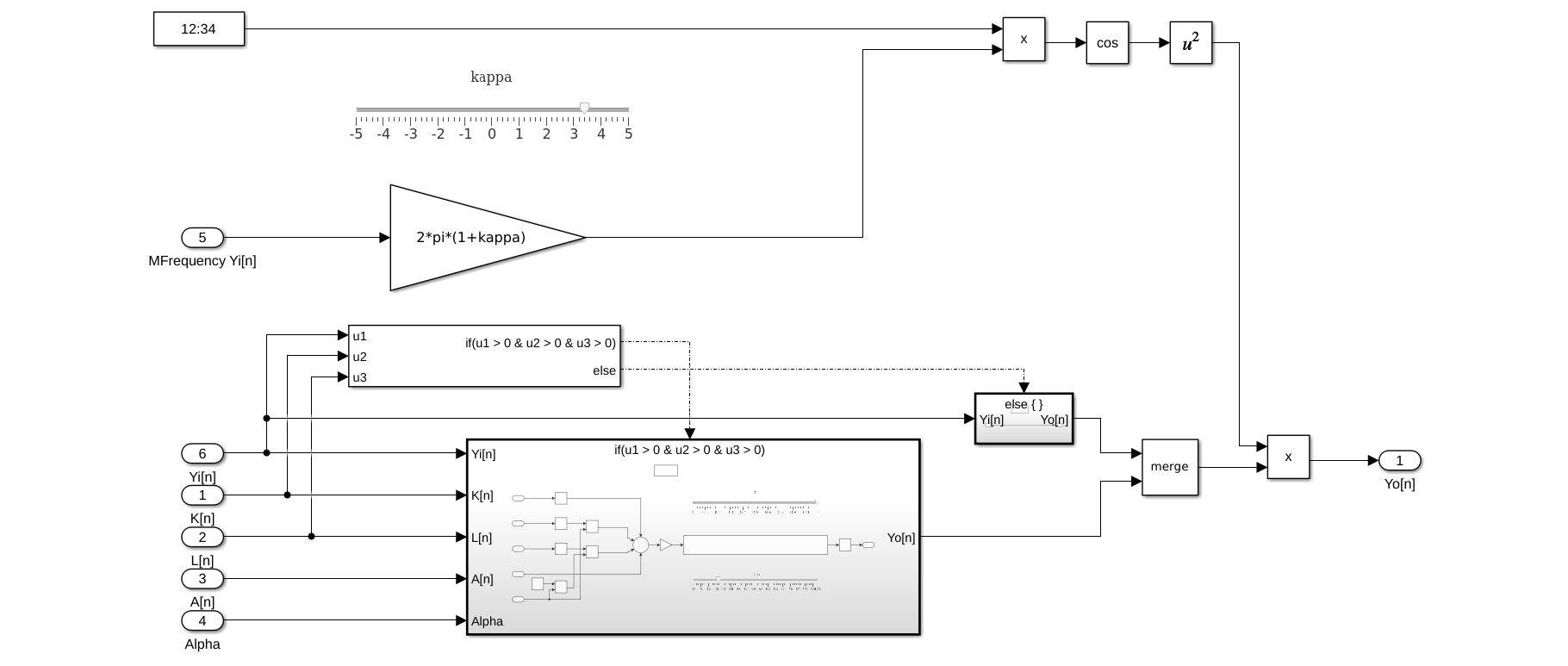}
                             \caption{Automated Task Model - II.}
                             \label{fig:AUTTASKII}
                         \end{minipage}
              \end{figure}

We filtered out signals per activity from each path. We simulated the  ``A\_PREACCEPTED'' task. Between 2012 and 2017, an automated system took the place of this job. See figure \ref{fig:AUTTASKI} and \ref{fig:AUTTASKII}. The activity's simulation results with $\kappa$ and $kappa_{wo}$ are displayed in table \ref{tab:AVGSIMT2012}.

              \begin{table}[!htb]
                \centering
                    \begin{tabular}{|c|c|c|}
                     \hline
                        Path name & $f_m [Hz]$ & $f_m [Hz]$\\
                                  & ``A\_PREACCEPTED'' & ``A\_PREACCEPTED'' \\
                                  & $\kappa$   & $\kappa_{wo}$ \\
                        \hline 
                        A & \num{7.88e-02} & \num{8.73e-02} \\ \hline 
                        B & \num{7.41e-02} & \num{8.11e-02} \\ \hline 
                        C & \num{6.64e-02} & \num{7.30e-02} \\ \hline
                    \end{tabular}
                    \caption{Intervened task -average frequencies per simulated path}
                    \label{tab:AVGSIMT2012}
               \end{table}

\begin{figure}[!htb]
                      \centering
                            \begin{minipage}{.5\textwidth}
                                \centering
                                \includegraphics[scale=0.13]{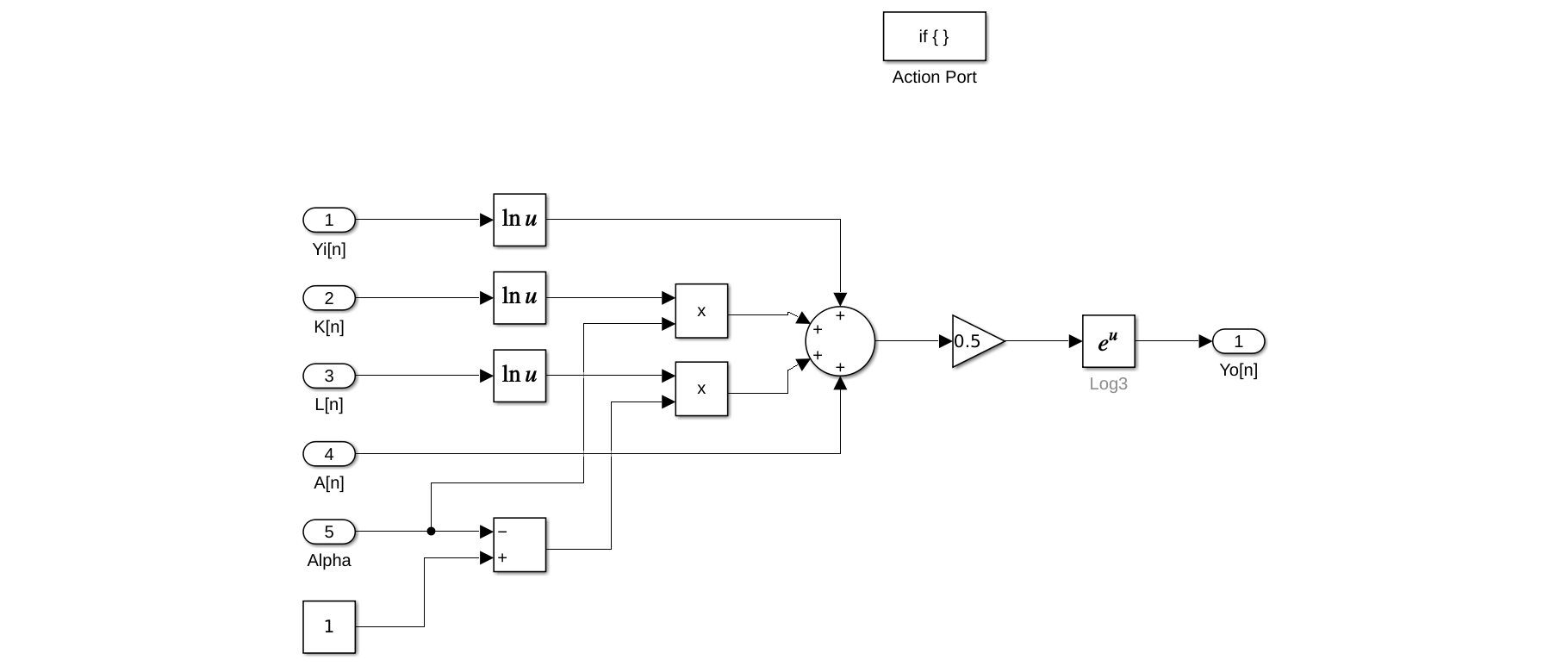}
                                    \caption{Non-Automated Task Model - I.}
                                    \label{fig:NAUTTASKI}
                                \end{minipage}%
                            \begin{minipage}{.6\textwidth}
                                   \centering
                                 \includegraphics[scale=0.12]{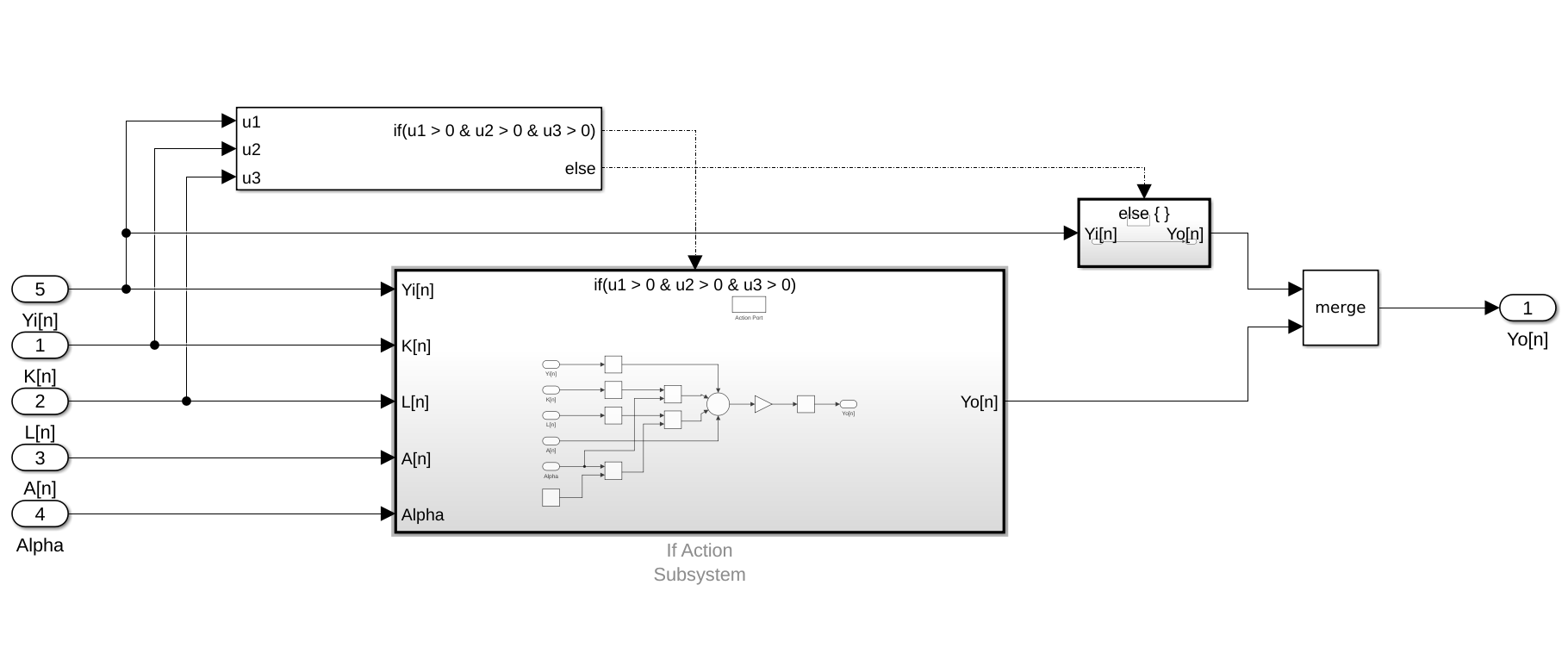}
                                    \caption{Non-Automated Task Model - II.}
                                    \label{fig:NAUTTASKII}
                            \end{minipage}
               \end{figure}

Our simulation model is shown in figure \ref{fig:AUTTASKII} from the outside. Figure \ref{fig:AUTTASKI} depicts the internal portion. This section's execution is synchronised by an "if" control. Figures \ref{fig:NAUTTASKI} and \ref{fig:NAUTTASKII} show the non-automated task model. The logic behind it is the same as automatic tasks.

\begin{figure}[!htbp]
                \centering 
                    {\includegraphics[scale=0.3]{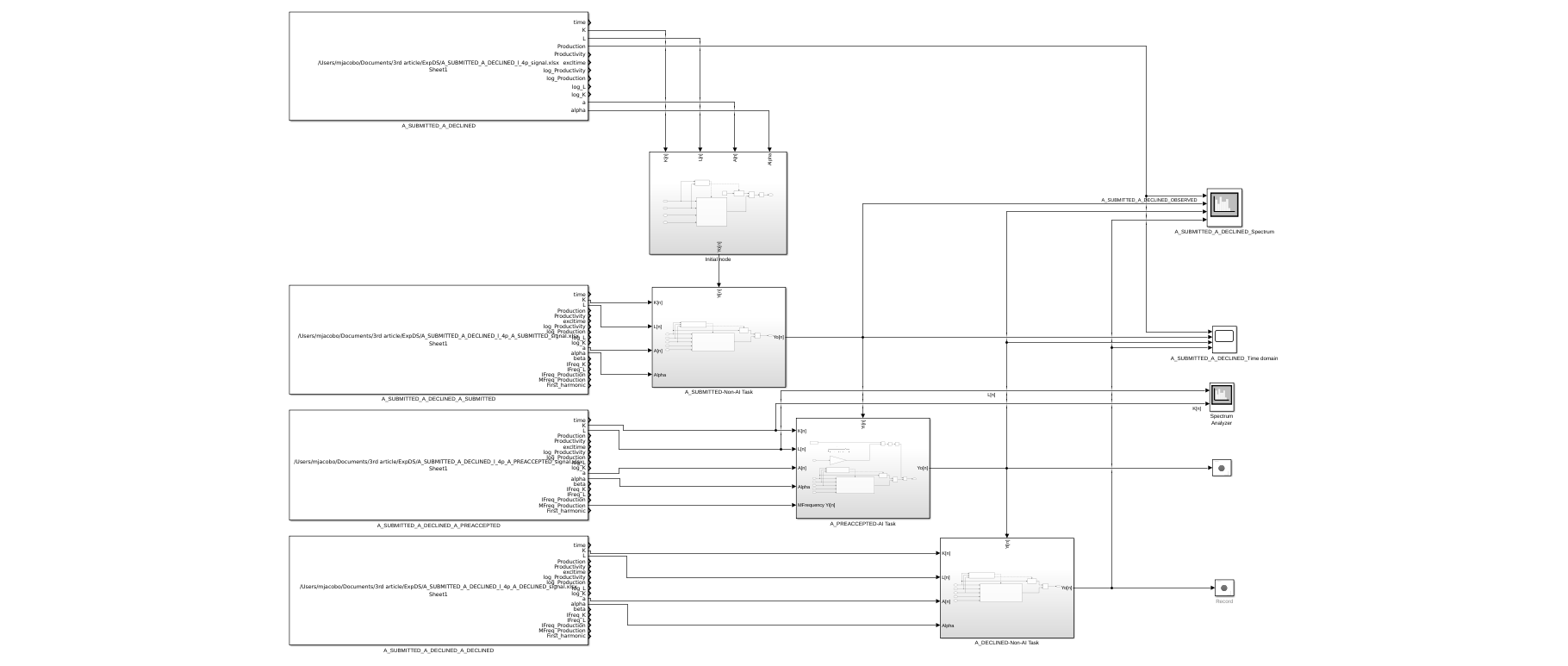}} 
                \caption{MATLAB\textsuperscript{\tiny\textregistered} - Simulation flowchart - A\_SUBMITTED\_A\_DECLINED.}
                \label{fig:SIMCRT1} 
            \end{figure}

Afterwards, we simulated the identified paths. Tables \ref{tab:AVGSIMFF2012} and \ref{tab:AVGSIMP2012} display the medium and fundamental frequencies for both kappa values, and figure \ref{fig:SIMCRT1} shows one illustration of the tested processes.
                \begin{table}[!htb]
                        \begin{minipage}{.5\linewidth}
                            \centering
                                \begin{tabular}{|c|c|c|}
                                    \hline
                                        Path name & $f_m [Hz]$ & $f_m [Hz]$\\
                                                  & $\kappa$   & $\kappa_{wo}$ \\
                                        \hline 
                                        A & \num{2.70e-03} & \num{2.90e-03} \\ \hline 
                                        B & \num{2.50e-03} & \num{3.00e-03} \\ \hline 
                                        C & \num{1.40e-03} & \num{1.60e-03} \\ \hline
                                \end{tabular}
                        \caption{Average frequencies per simulated path.}
                        \label{tab:AVGSIMFF2012}
                        \end{minipage}%
                        \begin{minipage}{.5\linewidth}
                            \centering
                                \begin{tabular}{|c|c|c|}
                                   \hline
                                       Path name & $f_0 [Hz]$ & $f_0 [Hz]$\\
                                                 & $\kappa$   & $\kappa_{wo}$ \\
                                       \hline 
                                       A & \num{2.39e-01} & \num{2.66e-01} \\ \hline 
                                       B & \num{2.22e-01} & \num{2.43e-01} \\ \hline 
                                       C & \num{1.97e-01} & \num{2.17e-01} \\ \hline
                                \end{tabular}
                        \caption{Fundamental frequencies per simulated path.}
                        \label{tab:AVGSIMP2012}
                        \end{minipage} 
                \end{table}

\section{Discussion}
\begin{figure}[!htb]
                    \centering
                        \begin{minipage}{.5\textwidth}
                            \centering
                            \includegraphics[scale=0.2]
                            {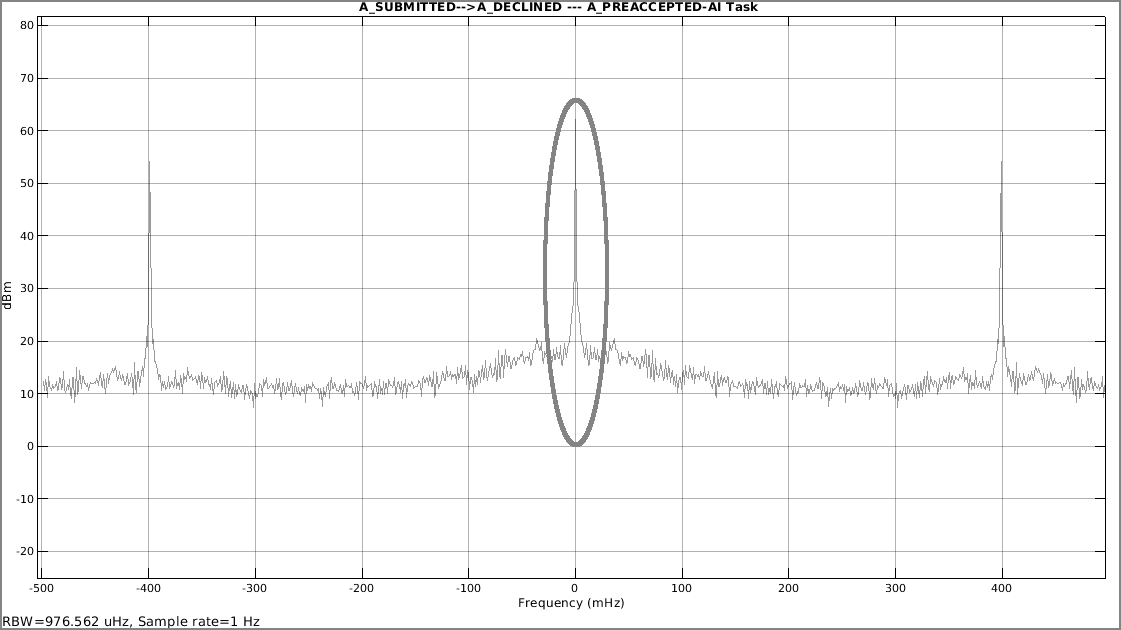} 
                            \caption{$A_a(t)$.}
                            \label{fig:SIMAT1} 
                         \end{minipage}%
                         \begin{minipage}{.5\textwidth}
                             \centering
                             \includegraphics[scale=0.2]                    {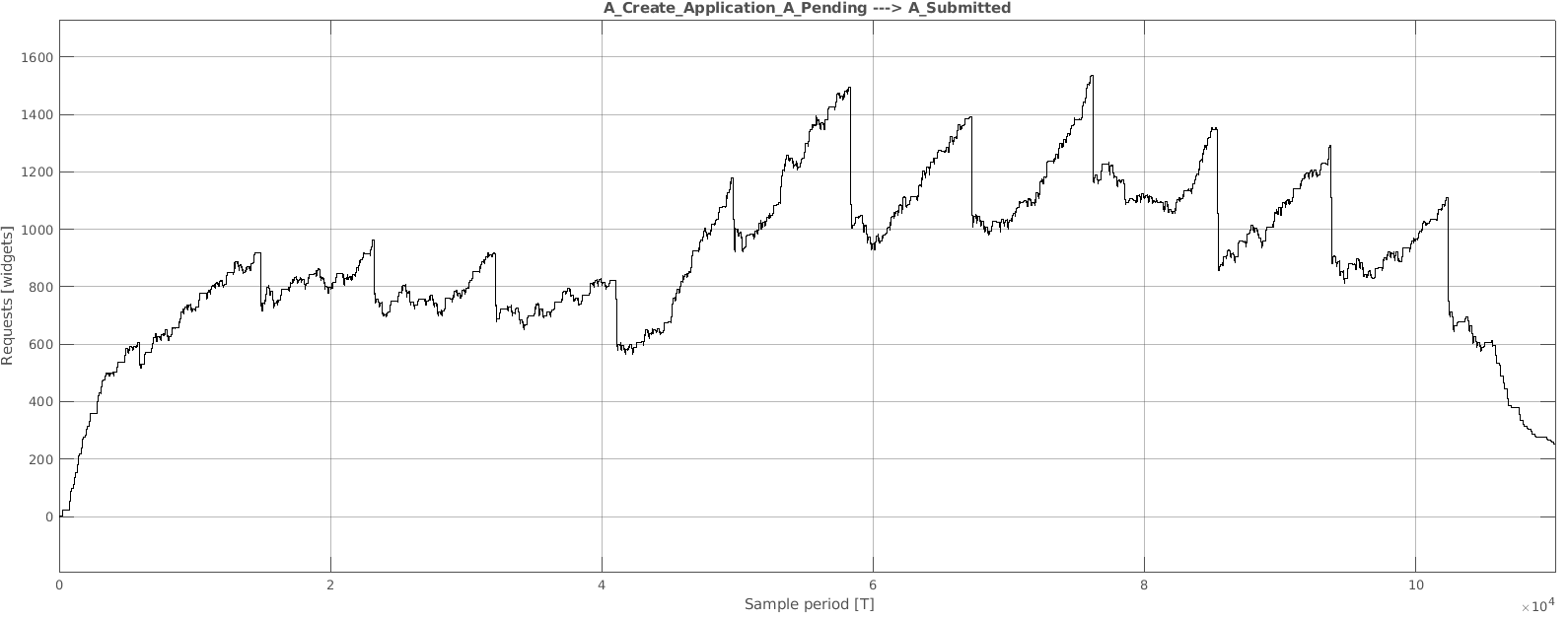} 
                             \caption{Task - A\_Submitted.}
                             \label{fig:BP2017CRSUAS} 
                         \end{minipage}
              \end{figure}

The interpretation of productivity as frequency allowed us to employ signal processing theory to provide granular information on productivity variations due to BP automation. In this fashion, a frequency acceleration is equivalent to a productivity increment and vice versa. Moreover, frequency shares the same mathematical characteristics as productivity. LTI system models allowed us to conceive productivity as a composition of individual task productivity. We employed the concept of superposition to create an LTI network. This approach allowed us to map each task as an LTI system. In this fashion, BP would be an LTI composed of the superposition of other LTIs. This approach ensures model scalability. 
After analysing public logs on an insurance process from two different years, we identified a frequency acceleration in the tasks that experienced automation. This finding confirmed our hypothesis.
On some occasions, automation generates new tasks. They add a burden that inhibits BP's frequency. The combined result is a decrease in the BP's productivity of the BP. Thus, there is a link between manual chores and low-frequency activities. This behaviour suggests the integration of new manual steps. Therefore, we can detect these deviations. 
After filtering new manual tasks, we found that BP performance improved.
We also calculated the values of the Cobb-Douglas equations: $A_a(t)$ and $\alpha$. $A_a(t)$ determines the energy amount at $\omega$=0 or potential energy.  This value indicates the portion of latent resources we employ to execute a task. $A_a(t)$ values are in logarithmic scale; thus, negative values represent weights between 0 and 1. See figure \ref{fig:SIMAT1}. On the other hand, $\alpha$ indicates the influence of capital, $K$, on the production equation. The $(1-\alpha)$ term is the labour weight.
Our work showed that BP's productivity tracks labour harmonic frequencies. This phenomenon reveals that automation accelerates BP frequency in the labour direction. The weights assigned in the Cobb-Douglas equation explain this. Hence, a change in labour frequency may produce the same effect as automation in the BP task.

              \begin{figure}[!htb]
                    \centering
                        \begin{minipage}{.5\textwidth}
                            \centering
                            \includegraphics[scale=0.19]
                            {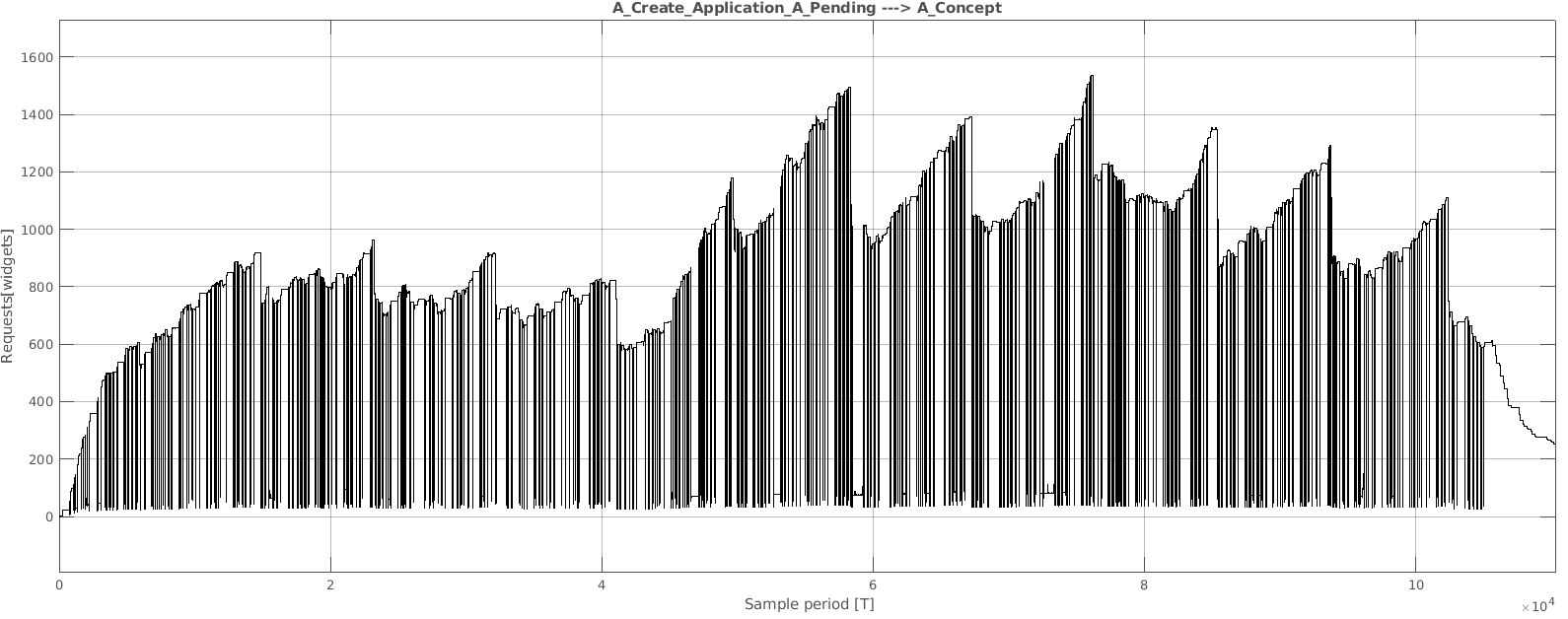} 
                            \caption{Task - A\_Concept.}
                            \label{fig:BP2017CRSUAC} 
                         \end{minipage}%
                         \begin{minipage}{.5\textwidth}
                             \centering
                             \includegraphics[scale=0.2]                    {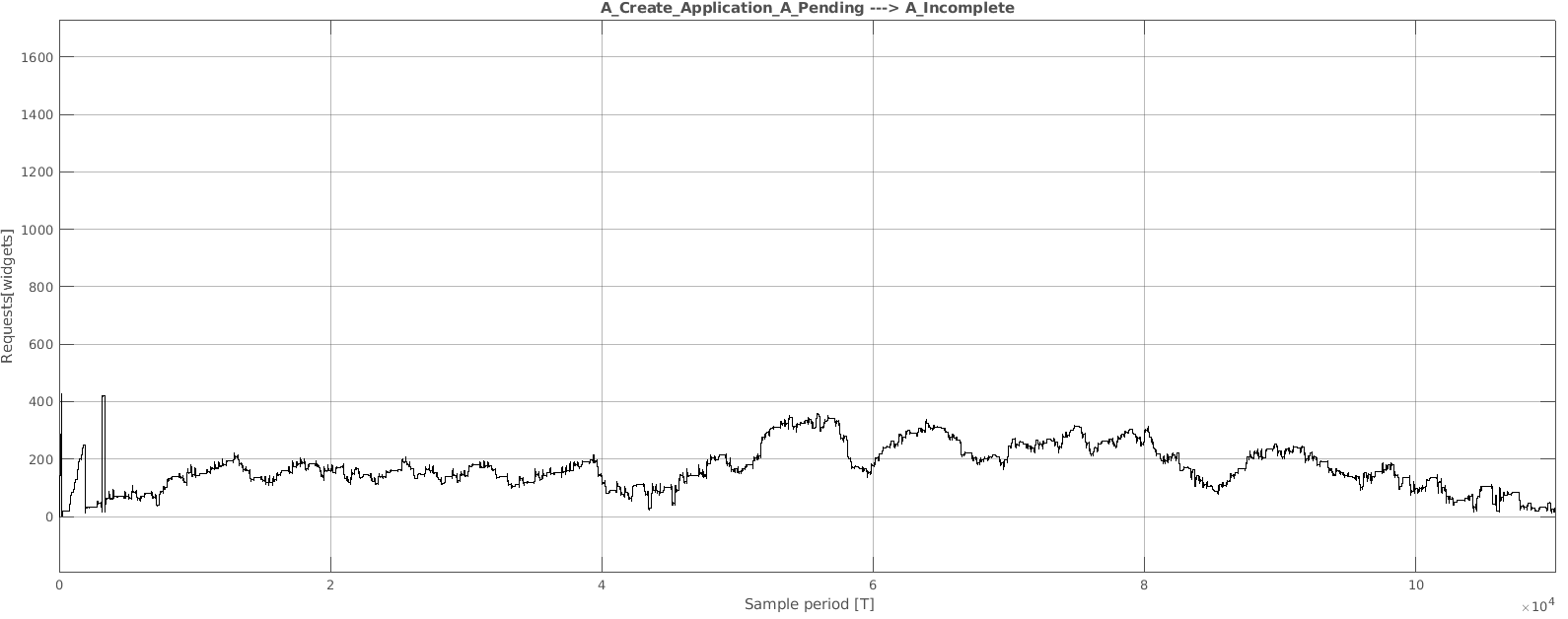} 
                             \caption{Task - A\_Incomplete.}
                             \label{fig:BP2017CRSUAI} 
                         \end{minipage}
              \end{figure}

Furthermore, we determined $\kappa$ values for the logs. Table \ref{tab:KAPPAV1217}  contains values of the total tasks and the filtered-out ones. This activity classification accounted for execution time to recognise automated chores with the analysis reports.

This taxonomy may introduce noise to our analysis. These data sets are open and don't contain specific details. Our model considered metrics as a control element. Therefore, we integrated the response on frequency as a mechanism to obtain optimal BP production quality. Parameters $\tau$ and $R_o$ helped us to identify the optimal quality value and its effect on the BP frequency answer. Additionally, these parameters may help to benchmark future automation. Our study found that $\tau > 20$ and $R_o  \approx 1$ provide a stability measurement band. Therefore, we can say that $\omega_r$ varies depending on these factors.
           
We observed a difference between the simulated productivity per path and the values registered in the logs. Further analysis identified that 2017's process had a new activity. This new job interfered with BP's productivity. BP's $f_m$ is smaller compared to simulated results. Figures \ref{fig:BP2017CRSUAS},  \ref{fig:BP2017CRSUAC}, and \ref{fig:BP2017CRSUAI} show the evolution of production throughout the BP. Figure \ref{fig:BP2017CRSUAC} shows the improvements in productivity due to automation. On the other hand, figure \ref{fig:BP2017CRSUAI} shows how the new task softens the frequency. As a result, 2017's BP declined its productivity due to automation.

We inferred that the ``A\_Incomplete"  task is manual since it generates a low frequency, $f_m$, that interferes with BP's automation profits due to the superposition principle. 
Following this analysis, it is possible to identify BP abnormalities or unexpected effects. Simulation considered no change in BP task inputs; therefore, we expected similar results.
Also, we introduced new techniques to productivity analysis. LTI system approach allowed us to scale up or down or review. Moreover,  Amplitude Modulation allowed us to accelerate the BP task's frequency according to a specific performance requirement. However,  carrier signal design might depend upon other variables.
Our design is also modular. We can simulate a significant portion of BP from different industries. This detailed information provides a piece of knowledge to resolve Solow's Paradox.
As further research, we recommend analysing the model with another modulation technique. We expect a model more resilient to noise. Moreover, an examination of ``Q'' value changes, from \ref{eq:MULTFPRODC1}, may help to determine the limits of the model causality. An optimum  ``Q'' value may improve resource employment when the system is idle. As the last point, the new metrics integration may change the value of $\omega_r$. It also may include new parameters integration that helps to control and optimise a better prediction of BP/BP-task productivity.

\section{Conclusions}
Our findings demonstrate that detailed information on BP and task activity levels is possible. Our model provides comprehensive data on how task automation affects productivity. We can monitor changes at a particular time and generalise this behaviour in the upcoming production cycle (RO1). The analysis in the frequency domain showed us that constant production is not the most efficient manner of transferring the value of capital and labour to BP. A still production implies that the system demands significant funds and work. A low frequency indicates elevated costs and resources. At the same time, high frequency represents lower costs but may lead the  BP  to unpredictable/unstable states (RO2). We integrated the accuracy metric as part of our model. This indicator allowed us to control the  BP task quality due to the introduction of automation. We considered that metrics act as negative feedback loops that enable us to amend the task/BP output. We defined two parameters, $\tau$ and $R_o$, to identify suitable frequency ranges to ensure task/BP stability according to firm requirements(RO3). Frequency analysis allowed us to detach our data from the time dimension. We considered the full logs as one production cycle. Thus, we directly compared the two files, 2012 and 2017. Besides, we can detect irregular production variations in other tasks due to automation. We can evaluate BP behaviour under current conditions. We can even notice whether an activity is already automated (RO4). Furthermore, we identified the resonant frequency, $\omega_r$, of our model. This metric depends on the weights of $\tau$ and $R_o$. We can consider this value as a limit and a criterion to define BP's low or high frequency. We presumed that a value from zero to  $\omega_r/2$ is low. Otherwise, it is a high one. A band around $\omega_r/2$ might be an optimal value for this metric. $\omega_r$ depends on the metric and the parameter values. Despite being calculated for the automated task, $\omega_r$ might also apply to non-automated activities (RO5).

\section*{Acknowledgments}
This work was supported by the Science and Technology Council of Mexico (Consejo Nacional de Ciencia y Tecnología, CONACYT, https://www.conacyt.gob.mx).

\bibliographystyle{unsrt}  
\bibliography{references}

\end{document}